\documentclass{article}
\usepackage{amsmath,graphicx,mlspconf,amsthm,amssymb}
\usepackage{algorithm}
\usepackage{amsmath}
\usepackage{url}
\usepackage{verbatim}

\DeclareMathOperator*{\Minimize}{Minimize}

\usepackage{algorithm}
\usepackage{algorithmic}

\copyrightnotice{Submitted to MLSP 2023}

\copyrightnotice{Submitted to MLSP 2023}

\copyrightnotice{Submitted to MLSP 2023}

\copyrightnotice{Submitted to MLSP 2023}

\toappear{}

\DeclareMathOperator{\rank}{rank}
\DeclareMathOperator{\argmin}{argmin}

\usepackage[usenames,dvipsnames]{color}

\definecolor{brightpink}{rgb}{1.0, 0.0, 0.5}

\newcommand{\ngi}[1]{{{\color{black} #1}}}

\title{Accelerated Algorithms for Nonlinear Matrix Decomposition \\ with the ReLU function}
\name{Giovanni Seraghiti$^\dagger$, 
Atharva Awari$^\ddagger$, 
Arnaud Vandaele$^\ddagger$, 
Margherita Porcelli$^\dagger$, 
Nicolas Gillis$^\ddagger$\thanks{NG acknowledges the support by the F.R.S.-FNRS and the FWO (EOS Project O005318F-RG47),  by the F.R.S.-FNRS under the PDR project T.0097.22, 
and by the Francqui Foundation.}} 
\address{$^\dagger$ University of Bologna \\ 
$^\ddagger$ University of Mons, Rue de Houdain 9, 7000 Mons}

\begin{document}

\maketitle

\begin{abstract}
In this paper, we study the following nonlinear matrix decomposition (NMD) problem: given a sparse nonnegative matrix $X$, find a low-rank matrix $\Theta$ such that $X \approx f(\Theta)$, where $f$ is an element-wise nonlinear function. We focus on the case where $f(\cdot) = \max(0, \cdot)$, the rectified unit (ReLU) non-linear activation. We refer to the corresponding problem as ReLU-NMD. We first provide a brief overview of the existing approaches that were developed to tackle ReLU-NMD. Then we introduce two new algorithms: (1) aggressive accelerated NMD (A-NMD) which uses an adaptive Nesterov extrapolation to accelerate an existing algorithm, and (2) three-block NMD (3B-NMD) which parametrizes $\Theta = WH$ and leads to a significant reduction in the computational cost.  We also propose an effective initialization strategy based on the nuclear norm as a proxy for the rank function. We illustrate the effectiveness of the proposed algorithms (available on gitlab) on synthetic and real-world data sets. 
\end{abstract}
\section{Introduction}
\label{sec:intro}

Low-rank matrix approximations are widely used in many fields such as data analysis and machine learning. When dealing with a large amount of data, stored in a matrix $X$, performing dimensionality reduction by approximating $X$ by a low-rank matrix is frequent in applications such as data compression,  interpretation, and visualization.  
In general, one wants to find a low-rank matrix $\Theta$ such that  $X \approx \Theta$. Some well-known examples of this type of approximation are the truncated singular value decomposition (TSVD)~\cite{eckart1936approximation}, and nonnegative matrix factorization (NMF)~\cite{gillis2020nonnegative}. 
In this context, there has been a recent emergence of nonlinear matrix decomposition (NMD), via its close connection with neural networks~\cite{saul2022nonlinear}. NMD looks for a low-rank matrix $\Theta$ such that $X \approx f(\Theta)$, where $f$ is an elementwise nonlinear function; of particular interest is the case where $f$ is the ReLU function, that is,  $f(\cdot)=\max(0,\cdot)$, which is often used as an activation function in hidden layers of  neural networks. 

The NMD problem we consider in this paper is the following: Given $X \in \mathbb{R}^{m \times n}$ and $r < \min(m,n)$, solve  
\begin{equation}
    \min_{\Theta \in \mathbb{R}^{m \times n}} \lVert X- \max (0, \Theta) \rVert_F^2 
    \; \text{ such that } \; \rank(\Theta)=r. 
    \label{eq:orig_nmd}
\end{equation}
We will refer to this problem as ReLU-NMD. 
ReLU-NMD makes sense only if $X$ is nonnegative, since \mbox{$\max (0, \Theta) \geq 0$}. Moreover, $X$ should be relatively sparse for ReLU-NMD to provide advantages compares to the TSVD: if $X$ has mostly positive entries, the solution of ReLU-NMD will be similar to that of the TSVD, since $\Theta$ will need to contain mostly positive entries. 
\ngi{For example, Saul showed that the identity matrix of any dimension can be exactly recovered with a rank-3 ReLU-NMD~\cite{saul2022nonlinear}. The reason is that  $\max(0,\Theta)$ can be full rank although $\Theta$ has low rank; see Section~\ref{sec:exp} for other examples.} 

The objective function in (\ref{eq:orig_nmd}) is neither differentiable nor convex, 
and the nonlinearity arising from the ReLU function makes the problem difficult to solve using a direct approach.  
ReLU-NMD has been recently investigated by Saul~\cite{saul2022nonlinear}, where he introduced 
another formulation for ReLU-NMD 
as the following latent variable model: 
\begin{equation}
     \min_{Z,\Theta} \lVert Z- \Theta \rVert_F^2 \quad 
     \mbox{such that} \quad \begin{cases} \rank(\Theta)=r, \\
    \max(0,Z)=X.
    \end{cases}
    \label{eq:lat_nmd}
\end{equation}
The main advantage of this new formulation is that the additional latent variable $Z$ allows one to move the nonlinearity from the objective function to the constraints, opening the possibility of exploring new solution strategies. 

In~\cite{saul2022nonlinear}, Saul presents two algorithms to solve~\eqref{eq:lat_nmd}: a naive algorithm, and an Expectation-Maximization (EM) algorithm. In a nutshell, they are both alternating minimization algorithms with respect to the two variables $Z$ and $\Theta$, and rely at each iteration on the computation of a rank-$r$ TSVD of an $m$-by-$n$ matrix; see Section~\ref{sec:previous} for more details. 
Furthermore, in \cite{saulgeometrical}, Saul uses a momentum acceleration step with fixed extrapolation parameter  to accelerate the convergence of the algorithms.

\paragraph*{Outline and contribution} In this paper, we provide new effective algorithms for ReLU-NMD. 
In Section~\ref{sec:previous}, we summarize the previous works on this problem, mostly that of Saul. 
In Section~\ref{sec:new}, we first include an adaptive momentum parameter estimation in the naive algorithm of Saul with the aim of  optimally self-tuning the parameter along the iterations.   
Then we introduce the three-block NMD (3B-NMD) algorithm, which exploits the parametrization $\Theta=WH$, and then uses an accelerated block coordinate method to update $W$, $H$ and $Z$. The main advantages of 3B-NMD is that it avoids the computation of a rank-$r$ TSVD at each iteration by solving instead least squares problems, reducing the computational cost from $O(mnr^2)$ to $O(mnr)$ operations.  
In Section~\ref{sec:nucnorm-init}, we explain how the nuclear norm can be used to provide a good initialization to ReLU-NMD. 
Finally, we illustrate on synthetic and real-world data sets the effectiveness of the proposed algorithms compared to the state of the art in Section~\ref{sec:exp}. 

\section{Previous algorithms by Saul}
\label{sec:previous}

In this section, we briefly recall the two main algorithms proposed by Saul; the first one will be the starting point of our newly proposed algorithms in Section~\ref{sec:new}. 

\paragraph*{Naive algorithm and extrapolation} A simple algorithm to tackle the reformulation~\eqref{eq:lat_nmd} is alternating optimization~\cite{saul2022nonlinear}.  
Let $I_+=\{ (i,j) \ | \ X_{ij}>0\}$ and $I_0=\{ (i,j) \ | \ X_{ij}=0\}$. At each iteration, $Z$ and $\Theta$ are computed alternatively: the optimal solution for $Z$, when $\Theta$ is fixed, is given by 
\begin{equation}
    Z_{ij} = \begin{cases}
         X_{ij}  & \text{ if } \quad (i,j) \in I_+,\\
         \min(0, \Theta_{ij})  & \text{ if } \quad (i,j) \in I_0.
    \end{cases}
    \label{eq:Z_naive}
\end{equation} 
The optimal solution for $\Theta$ when $Z$ is fixed is the rank-$r$ TSVD of $Z$. We will refer to this algorithm as Naive. 

In~\cite{saulgeometrical}, Saul accelerates the above simple scheme by an additional momentum term on the update of $Z$ with fixed momentum parameter~\cite{polyak1964some}. Denoting by $Z^{k+1}$, the $k+1$-th iterate, after $Z^{k+1}$ is computed, it is updated as  
\begin{equation}
    Z^{k+1} \leftarrow Z^{k+1}+\alpha(Z^k-Z^{k-1}),  
    \label{eq:pol_mom}
\end{equation}
where $\alpha \in (0,1)$ and it the experiments we set $\alpha=0.7$. We will refer to the accelerated Naive algorithm as A-Naive. 

\paragraph*{Expectation-Maximization (EM-NMD)} In~\cite{saul2022nonlinear}, Saul proposes a second more sophisticated EM algorithm, where the matrix $\Theta$ parameterizes the following Gaussian latent variable model for the data $X$: for all $i,j$, 
    $\tilde Z_{ij} \sim \mathcal{N}(\Theta_{ij}, \sigma^2)$. 
It is assumed that the observation, $Z$, is a sample of $\tilde{Z}$, and the matrix $X$ is obtained from the elementwise nonlinear mapping of $Z$: $X = \max(0,Z)$. 
The model is then estimated by maximizing the likelihood of the observation $X$ in terms of the matrix $\Theta$ and variance $\sigma^2$. The overall log-likelihood under this model is given by 
\begin{equation}
     \log P(X| \Theta, \sigma^2) = \Sigma_{ij} \log P(X_{ij}| \Theta _{ij}, \sigma^2) . 
\end{equation}
The parameters $\Theta$ and $\sigma^2$ are estimated by maximizing this sum.  
To do this, EM is used: it alternates between two steps: the E-step computes the posterior means and variances of the model latent variables, and the M-step uses these posterior statistics to re-estimate the model parameters. These steps are rather technical, and we refer the interested reader to~\cite{saul2022nonlinear} for more details. Note that the M-step requires the computation of a rank-$r$ TSVD, as in the naive approach, and hence requires $O(mnr^2)$ operations. 
In \cite{saulgeometrical} a momentum step as in \eqref{eq:pol_mom} is added to the algorithm, we will refer to it as A-EM.


\section{Two new algorithms for ReLU-NMD}
\label{sec:new}

Let us present our two new algorithms for ReLU-NMD. 

\subsection{Aggressive momentum algorithm (A-NMD)}
\label{ssec:aggr_mom}

In the naive algorithm, Saul used a Polyak-type extrapolation with a fixed momentum parameter and only extrapolated the variable $Z$; see Section~\ref{sec:previous}. 
Here we adopt a more aggressive Nesterov-type  extrapolation with a heuristic approach to tune this parameter, as we set
\begin{equation}
    Z^{k+1} \gets Z^{k+1}+ \beta_k(Z^{k+1}-Z^k).
    \label{eq:Z_nest}
\end{equation} 
 where $\beta_k$ is chosen adaptively. Also note that we use extrapolation for both variables, $Z$ and $\Theta$. 
The adaptive choice of the momentum parameter allows the algorithm to be less sensitive to that parameter, and adapt depending on the problem at hand. To do so, we follow the scheme in~\cite{ang2019accelerating} where it was used for NMF. 
In that scheme, the momentum parameter, $\beta_k$ at iteration $k$, is updated based on the decrease/increase of the objective function as follows. 
Fix the hyperparameters $1<\Bar{\gamma}<\gamma<\eta$. 
The momentum parameter is multiplied by $\gamma$ as long as the objective function is decreasing, unless it reaches the adaptive upper bound $\Bar{\beta}$. 
If the objective function decreases, we set $\Bar{\beta}=\min (1,\Bar{\gamma} \Bar{\beta})$, meaning that we increase $\Bar{\beta}$ by a factor $\Bar{\gamma}<\gamma$. On the contrary, if at iteration $k$ the error increases, the momentum parameter is divided by the factor $\eta$ and we update the upper bound $\Bar{\beta}$ as $\beta_{k-1}$, meaning that $\Bar{\beta}$ keeps track of the latest value of $\beta$ that allowed the decrease of the objective function. Algorithm~\ref{alg: naive} summarizes this strategy which we refer to as A-NMD. In the numerical experiments, we will use the parameters 
$\Bar{\gamma} = 1.05  
< 
\gamma = 1.1 
< 
\eta = 2.5$. 



\begin{algorithm}[ht!]
\caption{Aggressive momentum NMD (A-NMD)}
\begin{algorithmic}[1] 
\REQUIRE $X$, $Z^0$, $\Theta^0$, $r$, $1<\Bar{\gamma}<\gamma<\eta$, $\beta_0 \in (0,1)$,  maxit.  
\ENSURE A rank-$r$ matrix $\Theta$ s.t. $X\approx\max(0,\Theta)$. 
    \medskip  
\STATE Set $\Bar{\beta}=1$, $Z_{ij}^k = X_{ij}$ for $(i,j) \in I_+$ and for $k=0,1$. 

\FOR{$k=0,1,\dots,$ maxit} 
    \STATE $Z^{k+1}_{ij}=\min(0,\Theta^{k}_{ij})$ for 
    $(i,j) \in I_0$. 
    \STATE $Z^{k+1} \gets Z^{k+1}+\beta_k(Z^{k+1}-Z^{k}).$
    \STATE $[U,D,V] = \text{TSVD}(Z^{k+1}, r).$ 
    \STATE $\Theta^{k+1} = UDV^T.$ 
    \STATE $\Theta^{k+1} \gets\Theta^{k+1}+\beta_k(\Theta^{k+1}-\Theta^{k}).$
    \IF{$ \rVert X - \max(0,\Theta^{k+1}) \lVert_F < \rVert X - \max(0,\Theta^{k}) \lVert_F$ }
    \STATE $\beta_{k+1}=\min (\Bar{\beta},\gamma \beta_k),$ $\Bar{\beta}=\min (1,\Bar{\gamma} \Bar{\beta}).$
    \ELSE
    \STATE $\beta_{k+1}=\beta_k \backslash \eta$, 
    $\Bar{\beta}=\beta_{k-1}$, 
    \STATE $Z^{k+1}=Z^k$, 
    $\Theta^{k+1}=\Theta^k$. 
    \ENDIF
\ENDFOR
\STATE $\Theta = \Theta^{k+1}$. 
\end{algorithmic}  
\label{alg: naive}
\end{algorithm}

\subsection{Three-block NMD algorithm (3B-NMD)}
\label{ssec:three_blocks}

All the previously described algorithms require the computation of a rank-$r$ TSVD at each step. To avoid this relatively expensive step, we substitute $\Theta \in \mathbb{R}^{m \times n}$ by the product $WH$, where $W \in \mathbb{R}^{m \times r}$ and $H \in \mathbb{R}^{r \times n}$. 
Hence we reformulate~\eqref{eq:lat_nmd} as follows 
\begin{equation*}
     \min_{Z,W,H} \lVert Z - WH \rVert_F^2 \quad 
     \mbox{such that} \quad \max(0,Z)=X. 
\end{equation*} 
As for $\Theta$, the minimization subproblems for $W$ and $H$ have closed-form solutions; in fact, they can be obtained simply by solving matrix least square problems which require $O(mnr)$ operations, instead of the $O(mnr^2)$ operations for the TSVD. 
To accelerate this block-coordinate descent method scheme, we also use extrapolation, after the computation of $Z$ and of $\Theta = WH$; see Algorithm~\ref{alg: 3b_mom} (3B-NMD). 
\begin{algorithm}[ht!]
\caption{Momentum three-block NDM (3B-NMD)}
\begin{algorithmic}[1]
\REQUIRE $X$, $Z^0$, $W^0$, $H^0$,$r$, $\beta$, maxit.
\ENSURE Two matrices $W$ and $H$ s.t. $X \approx \max(0,WH).$
\STATE Set $Z_{ij}^k = X_{ij}$ for $(i,j) \in I_+$ and $k=0,1$. 
\FOR{$k=0,1,\dots$, maxit}
    \STATE $Z^{k+1}_{ij}=\min(0,\Theta^{k}_{ij})$ for 
    $(i,j) \in I_0$. 
    \STATE $Z^{k+1} \gets Z^{k+1}+\beta(Z^{k+1}-Z^{k}).$
    \STATE $W^{k+1} \gets$ $\argmin_W \lVert Z^{k+1} - W H^{k}  \rVert_F^2.$
    \STATE
    $H^{k+1} \gets \argmin_H \lVert Z^{k+1} - W^{k+1} H  \rVert_F^2.$
    \STATE $\Theta^{k+1} \gets W^{k+1} H^{k+1}$
    \STATE $\Theta^{k+1} \gets\Theta^{k+1}+\beta(\Theta^{k+1}-\Theta^{k}).$
\ENDFOR
\STATE $W =  W^{k+1}$, $H =  H^{k+1}$.  
\end{algorithmic}
\label{alg: 3b_mom}
\end{algorithm}

Note that 3B-NMD does not use an adaptive strategy for the momentum parameter $\beta$, because we have observed it is not as effective as in the naive case described in the previous section. This is a topic for further research. In the numerical experiments, we will use $\beta = 0.7$ which performs well in practice.


\section{INITIALIZATION WITH THE NUCLEAR NORM}
\label{sec:nucnorm-init}


In this section, we provide an initialization strategy for $\Theta$ using the nuclear norm.
The nuclear norm of a matrix $X$ is the sum  of its singular values, and is denoted $\|X\|_* = \sum_i \sigma_i(X)$. 
The nuclear norm has been used as a convex surrogate of the rank function, akin to the $\ell_1$ norm used as a convex surrogate for the $\ell_0$ norm; 
see~\cite{recht2010guaranteed} and the references therein. 

\ngi{Assuming there exists an exact rank-$r$ ReLU-NMD but we do not know the rank $r$, the rank identification problem can be reformulated as follows}: 
\begin{equation} \label{rankform} 
\min_{\Theta} \rank( \Theta ) 
\; \text{ such that }  \; 
X = \max(0, \Theta), 
\end{equation}
which is a hard problem in general. 
Replacing the rank with the nuclear norm, we obtain the following convex relaxation: 
\begin{align} 
\min_{\Theta} \| \Theta \|_* \; \text{ such that }  \; 
& \Theta_{ij} = X_{ij}  \text{ for } (i,j) \in I_+, \nonumber \\ 
& \Theta_{ij} \leq 0 \text{ for } (i,j) \in I_0. \label{nucnorm} 
\end{align}
\ngi{We are looking for $\Theta$ that matches the positive entries of $X$ while having the smallest possible nuclear norm, and hence, hopefully, a small rank.} 
To solve~\eqref{nucnorm}, we resort to a standard projected subgradient strategy~\cite{recht2010guaranteed}, updating $\Theta$ as follows 
\begin{equation*}
    \Theta_{k+1} = \Pi (\Theta_k - \alpha_k Y_k), \hspace{1cm} Y_k \in \partial \|\Theta_k\|_*, 
    \label{eq:proj_nuc}
\end{equation*}
where $\Pi(\cdot)$ is the projection onto the feasible set (which is easy to compute), and $\partial \|\Theta_k\|_*$ is a subgradient of the nuclear norm at $\Theta_k$, given by \cite{watson1992characterization} 
\begin{multline}
    \partial \| \Theta \|_* = \Big\{ UV^T + P : \text{$P$ and $\Theta$ have orthogonal row} \\\text{and column spaces, and } \|P\| \leq 1 \Big\}, 
    \label{eq:sub_nuc}
\end{multline} 
where $(U,\Sigma,V) \in 
\mathbb{R}^{m \times r} \times \mathbb{R}^{r \times r} \times \mathbb{R}^{n \times r}$ is a TSVD of $\Theta$ and $\|.\|$ is the operator norm (or induced 2-norm) of a matrix. In our implementation, we used $P = 0$. 

There are several possibilities for the choice of the stepsizes $\alpha_k$. A standard choice that guarantees convergence is to use a diminishing stepsize. 
However, we prefer to use a more aggressive  backtracking approach, since we only needed to perform a 
few iterations of 
the subgradient algorithm as it was sufficient to obtain a significant decrease in the nuclear norm. Also, the solution is only used for initialization and hence we do not need high accuracy. 
Moreover, the solution of~\eqref{nucnorm} is not guaranteed to be of rank smaller than $r$, and hence we use the rank-$r$ TSVD of the last iterate as an initialization for the ReLU-NMD algorithms presented in the previous sections. 
Note that it would be however rather interesting to explore conditions under which the solution of~\eqref{nucnorm} recovers that of~\eqref{rankform}, similarly as done in the affine rank minimization literature~\cite{recht2010guaranteed}.  

\section{Numerical experiments}
\label{sec:exp}

We compare the following algorithms for ReLU-NMD:  
 A-NMD (Algorithm~\ref{alg: naive}),  
 3B-NMD (Algorithm~\ref{alg: 3b_mom}), Naive-NMD, A-Naive-NMD, EM-NMD and A-EM by\footnote{We thank Lawrence Saul for providing us with his MATLAB codes.} Saul~\cite{saul2022nonlinear} (see Section~\ref{sec:previous}).  
 As a baseline, we will also report the result of the projection of the TSVD, that is, $\max(0,X_r)$ where $X_r$ is the rank-$r$ TSVD of $X$. 
 All tests are preformed using Matlab R2021b on a laptop Intel CORE i5-1135G7 @ 2.40GHz 8GB RAM. In the following, we perform experiments on synthetic data sets and the MNIST data set. \ngi{The code is available online from \url{https://gitlab.com/ngillis/ReLU-NMD}.} 




\subsection{Synthetic data}
\label{ssec:syn}

The matrix $X \in \mathbb{R}^{m \times n}$ is generated as $X=\max(0,WH)$, where the entries of $W \in \mathbb{R}^{m \times r}$ and $H \in \mathbb{R}^{r \times n}$ are generated from the normal distribution, that is, \texttt{W = randn(m,r)} and \texttt{H = randn(r,n)} in MATLAB. 
Since the probability for $X$ to have a positive entry is equal to that of having a negative entry (by symmetry), $X$ has, on average, 50\% of its entries equal to zero. 
All algorithms are stopped when 
\begin{equation}
\text{relative error} =\frac{\lVert X-\max(0,\Theta) \rVert_F}{\lVert X \rVert_F} \; \leq \; 10^{-4}, 
\label{eq:rel_err_std}
\end{equation}
where $\Theta$ is the current solution. 
Note that, given the non-convexity of ReLU-NMD, there is no guarantee that an algorithm converges to such a small relative error. However, for the synthetic data as generated above, this is always the case. 
It would be interesting to explain this behavior, e.g., maybe there are no spurious local minima with high probability, as for other non-convex problems~\cite{ge2017no}. 

\paragraph*{Initialization} Let us first validate the effectiveness of the nuclear norm initialization. 
To do so, we compare:  

\noindent $\bullet$ Random initialization where  $\Theta$ is a rank-$r$ matrix, generated in the same way as $X$, and which we scale optimally as follows $\Theta \leftarrow \alpha^* \Theta$ where 
\begin{equation*}
 \alpha^* =   \argmin_{\alpha} \| X - \alpha \ \max(0, \Theta)\|_F = \frac{\langle X, \max(0, \Theta) \rangle}{\| \max(0, \Theta) \|_F^2}. 
\end{equation*}  

\noindent $\bullet$ The initialization $\Theta$ taken as the rank-$r$ TSVD of $X$. 

\noindent $\bullet$ The nuclear norm minimization   
strategy; see Section \ref{sec:nucnorm-init}. We  perform 3 iterations of the projected subgradient method (which we observe is a good tradeoff between computational cost and reduction in the error). We initialize the nuclear norm approach using the random initialization and optimal scaling as described above.

Table~\ref{tab:nuclear_init_rand_tsvd} reports the relative errors of all three initializations for two values of $r$, with $m=n=500, 1000, 1500, 2000$. 
\begin{table}[h!]
\centering
\begin{tabular}{| c | c | c | c | c | c | c |} 
 \hline
         & \multicolumn{3}{c|}{$r=8$}
                    & \multicolumn{3}{c|}{$r=16$}
                                         \\
\hline 
 $m=n$  & rand & TSVD & $\|.\|_*$  & rand & TSVD & $\|.\|_*$\\
 \hline\hline
 $500$  & 0.95 &0.40 & \textbf{0.36} & 0.95 & 0.36 & \textbf{0.32} \\
 \hline
 $1000$ & 0.95 & 0.41 & \textbf{0.38} &  0.95 & 0.37 & \textbf{0.33}   \\
 \hline
 $1500$ & 0.95 & 0.41 & \textbf{0.38} & 0.95 & 0.38  & \textbf{0.33}    \\
 \hline
$2000$  & 0.95 & 0.41 & \textbf{0.38} &  0.95 & 0.38 &  \textbf{0.33}   \\
 \hline
\end{tabular}
\caption{Initial relative error of three initializations: random initialization 
and scaling, the rank-$r$ TSVD, and  the nuclear norm initialization described in Section~\ref{sec:nucnorm-init} ($\|.\|_*$).}
\label{tab:nuclear_init_rand_tsvd}
\end{table}  
We observe that the nuclear norm allows an initial solution with significantly smaller relative error: a reduction from a factor 2 to 3 compared to a random initialization, and about 10\% improvement compared to the TSVD. Hence, in the remaining of the paper, we will use the nuclear norm initialization for all algorithms in all experiments.


\paragraph*{Effectiveness of A-NMD}  
Let us first show the effectiveness of the adaptive strategy used in A-NMD to accelerate the naive scheme by Saul described in Section~\ref{sec:previous}. 
Table \ref{tab:naive_comp} reports the total time and iterations needed to reach a relative error as in (\ref{eq:rel_err_std}). We considered averaged values over 5 different synthetic matrices, with 5 different random initializations, post-processed using the nuclear norm algorithm. 

\begin{table}[h!]
\centering
\begin{tabular}{| c | c | c | c | c | c | c |} 
 \hline
   & \multicolumn{2}{c|}{Naive}
            & \multicolumn{2}{c|}{A-Naive}
                    & \multicolumn{2}{c|}{A-NMD}
                                         \\
\hline 
  Size & time & iter  & time& iter & time  &  iter   \\
 \hline\hline
 500 & 1.84 & 110  & 0.78& 44 & \textbf{0.57}  &  32  \\
 \hline
 1000  & 7.21 & 87  &2.85 & 33 &  \textbf{2.28} &  27   \\
 \hline
 1500  & 11.4 &  80 &4.40 & 29 & \textbf{3.71}  &  25    \\
 \hline
 2000  & 21.2 & 78  & 8.19& 28 &  \textbf{6.80} &   24   \\
 \hline
\end{tabular}
\caption{Average computational time needed to satisfy condition in (\ref{eq:rel_err_std}) on synthetic data with $r=32$.} 
\label{tab:naive_comp}
\end{table}
We observe that A-NMD outperforms Naive and A-Naive: it about 4 times faster than Naive, and 20\% faster, on average, than A-Naive. 
In the following experiments, we will therefore only compare the other algorithms with A-NMD.

\paragraph*{Effectiveness of 3B-NMD} 

Table~\ref{tab:size_comp} reports the total time and iterations needed to satisfy condition in (\ref{eq:rel_err_std}), for a fixed rank $r=32$, for the EM algorithm of Saul (EM-NMD), its accelerated variant (A-EM), as well as our proposed algorithms A-NMD and 3B-NMD.  Table \ref{tab:rank_comp} reports the same quantities for fixed dimensions   $m=n=1000$, but with different values of the rank, $r$.
\begin{table}[h!]
\centering
\begin{tabular}{| c | c | c | c | c | c | c | c | c |} 
 \hline
   & \multicolumn{2}{c|}{A-NMD}
            & \multicolumn{2}{c|}{3B-NMD}
                    & \multicolumn{2}{c|}{EM-NMD}
                        & \multicolumn{2}{c|}{A-EM}
                                         \\
\hline 
   \hspace{-0.2cm} Size \hspace{-0.2cm} & time & iter  & time& iter & time  &  iter & time  &  iter    \\
 \hline\hline
 \hspace{-0.2cm} 500 \hspace{-0.2cm}  & 0.64 & 33  &\textbf{0.08} & 23 & 2.2  &  101 & 1.0  &  43  \\
 \hline
 \hspace{-0.2cm} 1000 \hspace{-0.2cm}   &2.4  &  27 &\textbf{0.24} & 24 & 8.1   &  78& 3.9   &  36    \\
 \hline
 \hspace{-0.2cm} 1500 \hspace{-0.2cm}   &3.6  &  24 & \textbf{0.46} & 24 &12.8   &  70 &6.5   &  35    \\
 \hline
 \hspace{-0.2cm} 2000 \hspace{-0.2cm}  &6.7  &  25 & \textbf{0.81} & 24  &22.1   &  66 &11.6   &  34    \\
 \hline
\end{tabular}
\caption{Time needed to satisfy condition in (\ref{eq:rel_err_std}) for synthetic matrices of increasing dimension with $r=32$.}
\label{tab:size_comp}
\end{table}
\begin{table}[h!]
\centering
\begin{tabular}{| c | c | c | c | c | c | c | c | c |} 
 \hline
  & \multicolumn{2}{c|}{A-NMD}
            & \multicolumn{2}{c|}{3B-NMD}
                    & \multicolumn{2}{c|}{EM-NMD}
                        & \multicolumn{2}{c|}{A-EM}
                                         \\
\hline 
  $r$  & time & iter  & time& iter & time  &  iter & time  &  iter    \\
 \hline\hline
 8 & 2.0 & 33  &\textbf{0.22} & 22 & 10.1  &  97 & 4.4  &  42  \\
 \hline
 16  & 2.0  &  24 &\textbf{0.20} & 24 & 7.7   &  77& 3.5   &  36    \\
 \hline
 32  &2.2  &  23 & \textbf{0.22} & 24 &7.4   &  72 &3.6   &  33    \\
 \hline
 64  &2.5  &  23 & \textbf{0.25} & 25  &8.4   &  66 &3.7   &  32    \\
 \hline
\end{tabular}
\caption{Computational time needed to satisfy condition in (\ref{eq:rel_err_std}) when approximating synthetic data of fixed size $n=m=1000$ for different values of the rank, $r$.}
\label{tab:rank_comp}
\end{table} 
We observe that A-EM performs better than EM-NMD, as expected, and hence we will report only the results for A-EM in the section on real-world data sets. 
Then, we observe that A-NMD performs better than A-EM (almost twice faster in all cases), while 3B-NMD outperforms all other algorithms, being more than 10 times faster than A-EM.  

\subsection{MNIST data set} 
\label{ssec:sparse_data}

\begin{figure}[htb]
\begin{minipage}[b]{.48\linewidth}
  \centering
  \centerline{\includegraphics[width=4.5cm]{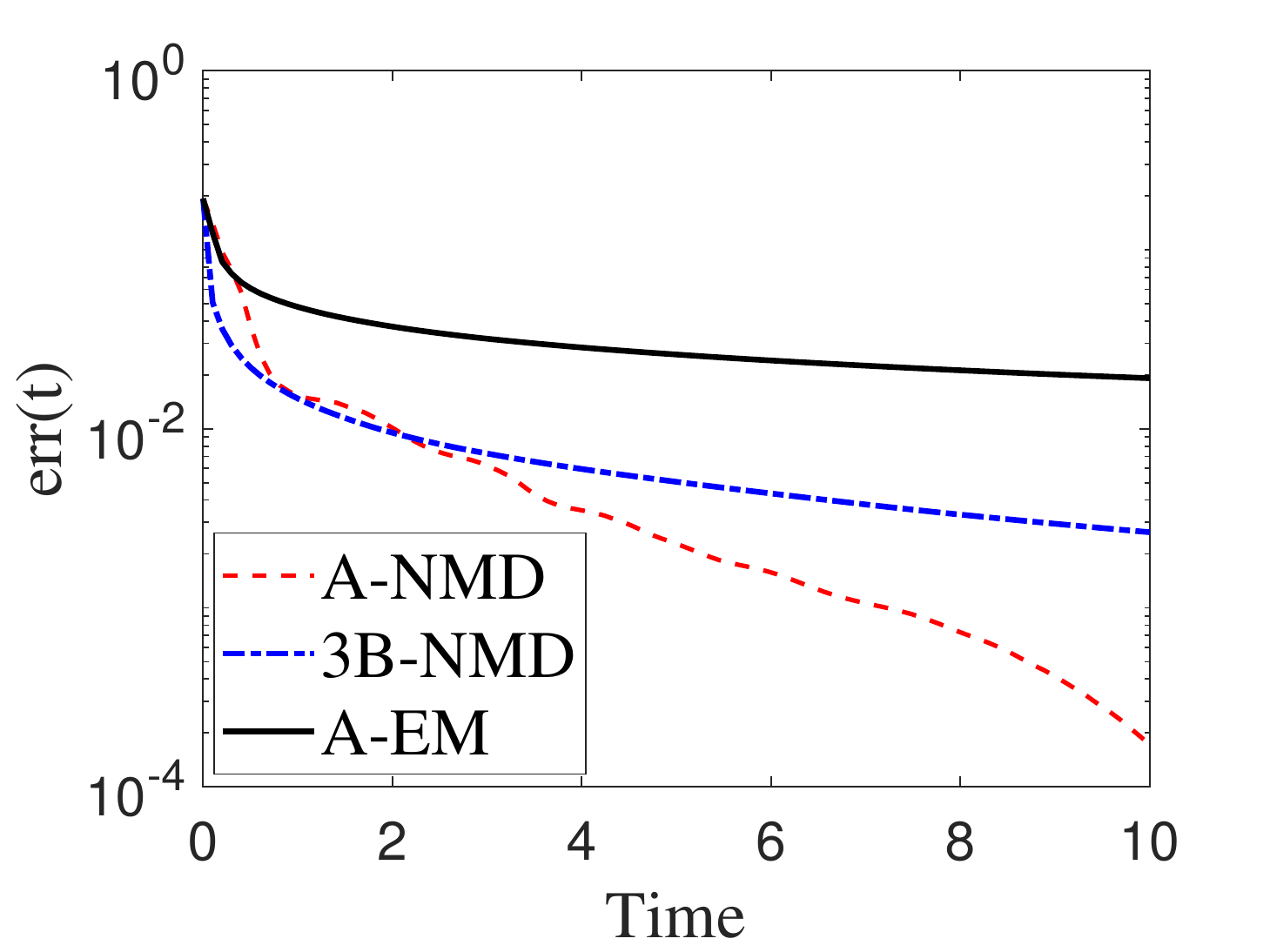}}
  \centerline{(a) $m=500$.}\medskip
\end{minipage}
\hfill
\begin{minipage}[b]{0.48\linewidth}
  \centering
  \centerline{\includegraphics[width=4.5cm]{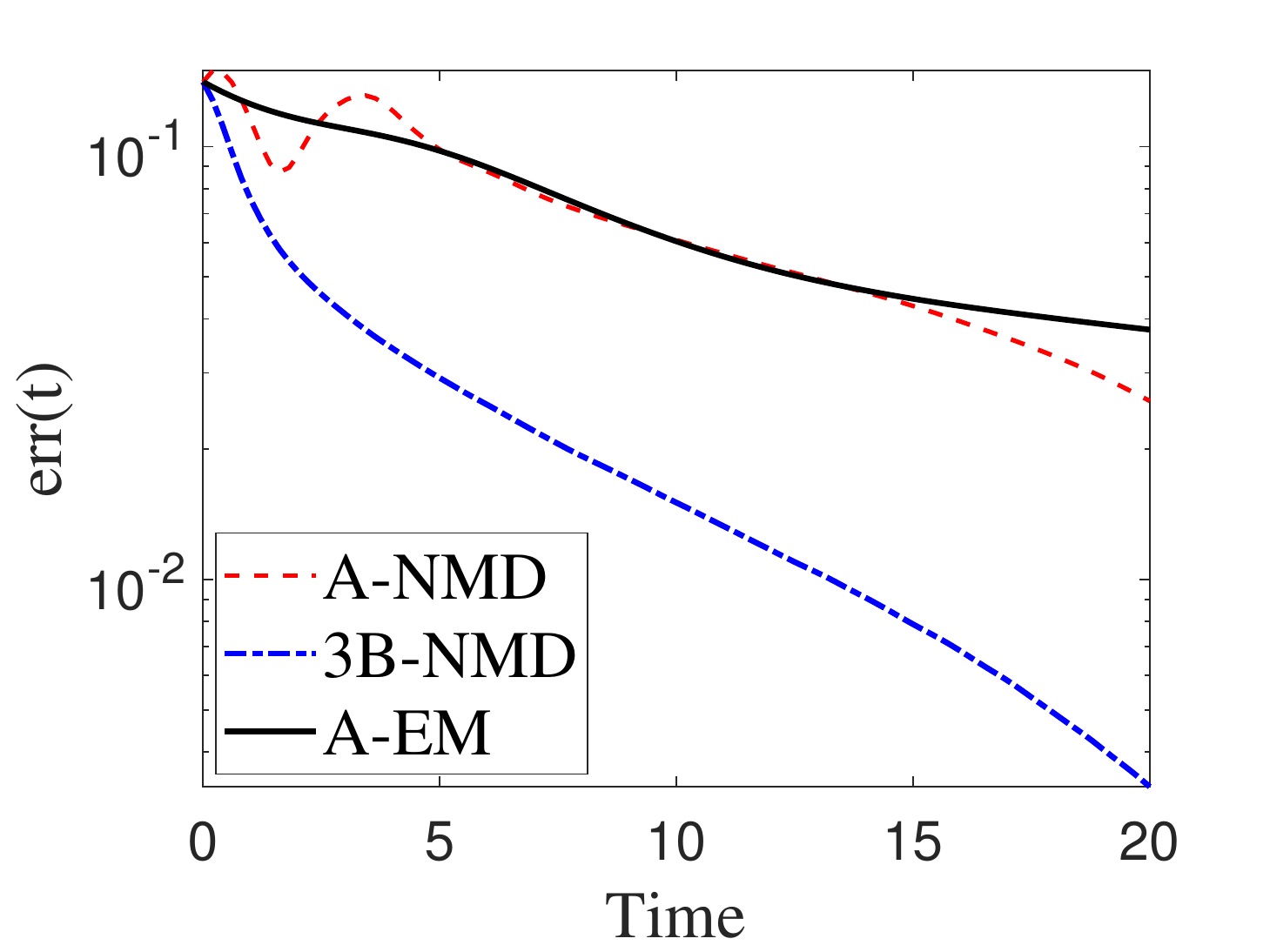}}
  \centerline{(b) $m=50000$.}\medskip
\end{minipage}
\caption{Average value of the error (\ref{eq:err}) of A-NMD, 3B-NMD and EM-NMD on small (a) and large (b) portion of the MNIST data set.}
\label{fig:avg_error}
\end{figure}

The experiments in this section are computed on $28 \times 28$ grayscale images of MNIST handwritten
digits~\cite{deng2012mnist}. We perform a first experiment with 500 images, and a second with 50000. 
To compare the ReLU-NMD algorithms, we will use the following quantity 
\begin{equation}
    err(t)=\frac{\lVert X-\max(0,\Theta (t)) \rVert_F}{\lVert X \rVert_F}-e_{min}, 
    \label{eq:err}
\end{equation}
where $\Theta(t)$ the solution at time $t$, and $e_{min}$ is the smallest relative error obtained by all algorithms within the allotted time. 
Since $err(t)$ converges to zero for the algorithm that computed the best solution, we can represent the error in log scale. 
Figure~\ref{fig:avg_error}~(a) displays the results on the MNIST data set with $r=32$ with 500 images (50 images of each digit), with a timelimit of 10 seconds. 
Although 3B-NMD converges initially faster, A-NMD eventually catches up and generates the best solution. As before, A-EM is outperformed. 


Figure~\ref{fig:avg_error}~(b) displays the results on the MNIST data set with $r=32$ with all the 50000 images, with a timelimit of 20 seconds. In this case, 3B-NMD converges initially faster and A-NMD does not have time to catch up. In any case, 3B-NMD and A-NMD both perform well, outperforming the state of the art algorithm A-EM. 


\begin{figure}[htb]
\begin{minipage}[b]{.48\linewidth}
  \centering
  \centerline{\includegraphics[width=4.5cm]{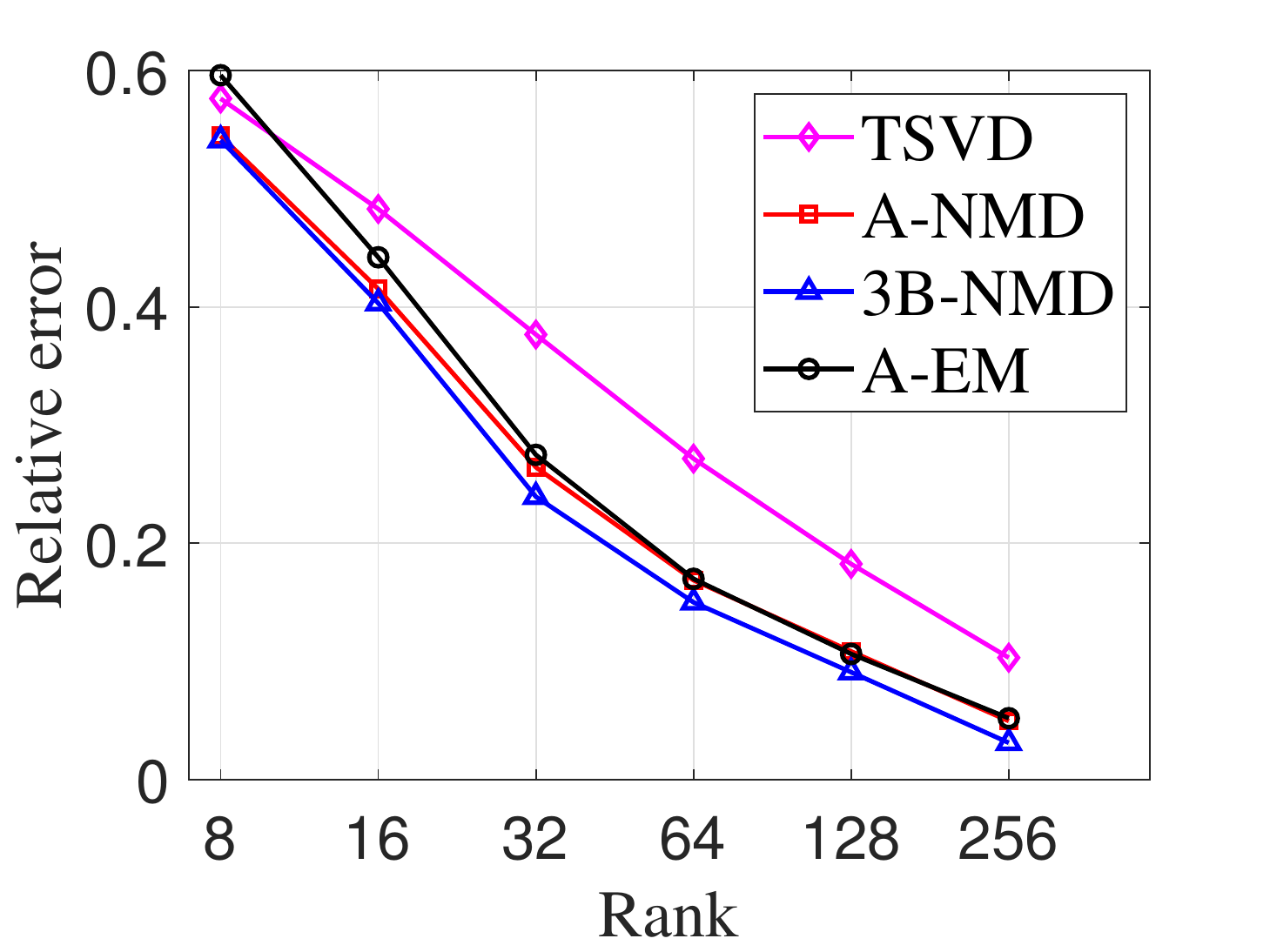}}
  \centerline{(a) Relative error.}\medskip
\end{minipage}
\hfill
\begin{minipage}[b]{0.48\linewidth}
  \centering
  \centerline{\includegraphics[width=4.5cm]{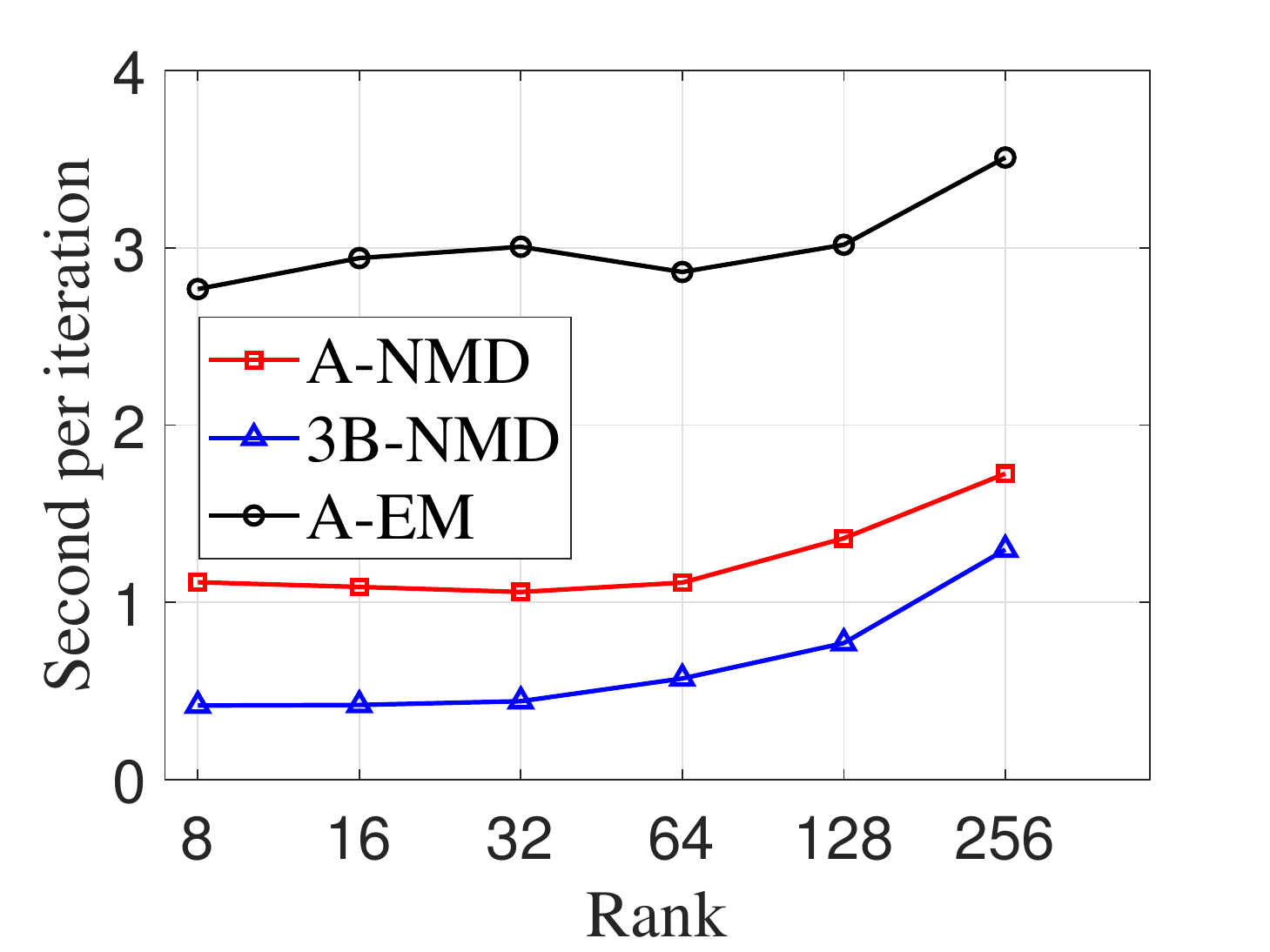}}
  \centerline{(b) Time (seconds).}\medskip
\end{minipage}
\caption{Final relative error on $m=50000$ images from MNIST dataset, after 20 seconds and average iteration time for increasing value of 
the rank $r$. }
\label{fig:rank_MNIST}
\end{figure}

Figure~\ref{fig:rank_MNIST} compares the algorithms with the baseline, TSVD,  in terms of relative error as the rank increases, and provides the average time per iteration, for 50000 images of MNIST and a runtime of 20 seconds. 
These results confirm the effectiveness of the 3B-NMD in order to deal with large data, the cost per iteration being smaller than that of A-NMD and A-EM. In addition, note that all the ReLU-NMD algorithms approximate the dataset with considerably higher accuracy than the TSVD, as expected.  

\subsection{Compression of sparse NMF basis} 
\label{ssec:NMF} 

\begin{figure}[htb]
\begin{minipage}[b]{.48\linewidth}
  \centering
  \centerline{\includegraphics[width=4.5cm]{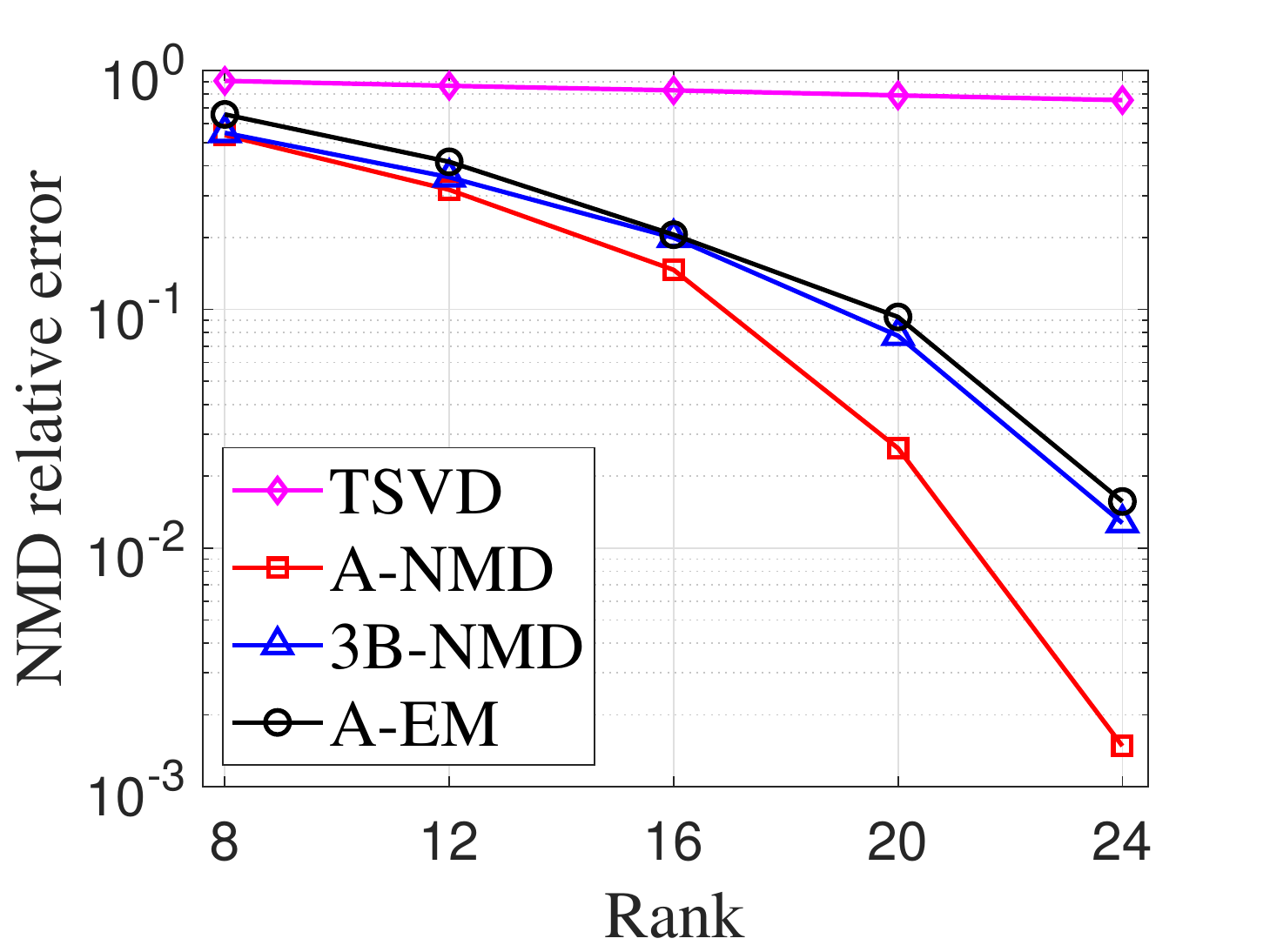}}
  \centerline{(a) Error on $U$.}\medskip
\end{minipage}
\hfill
\begin{minipage}[b]{0.48\linewidth}
  \centering
  \centerline{\includegraphics[width=4.5cm]{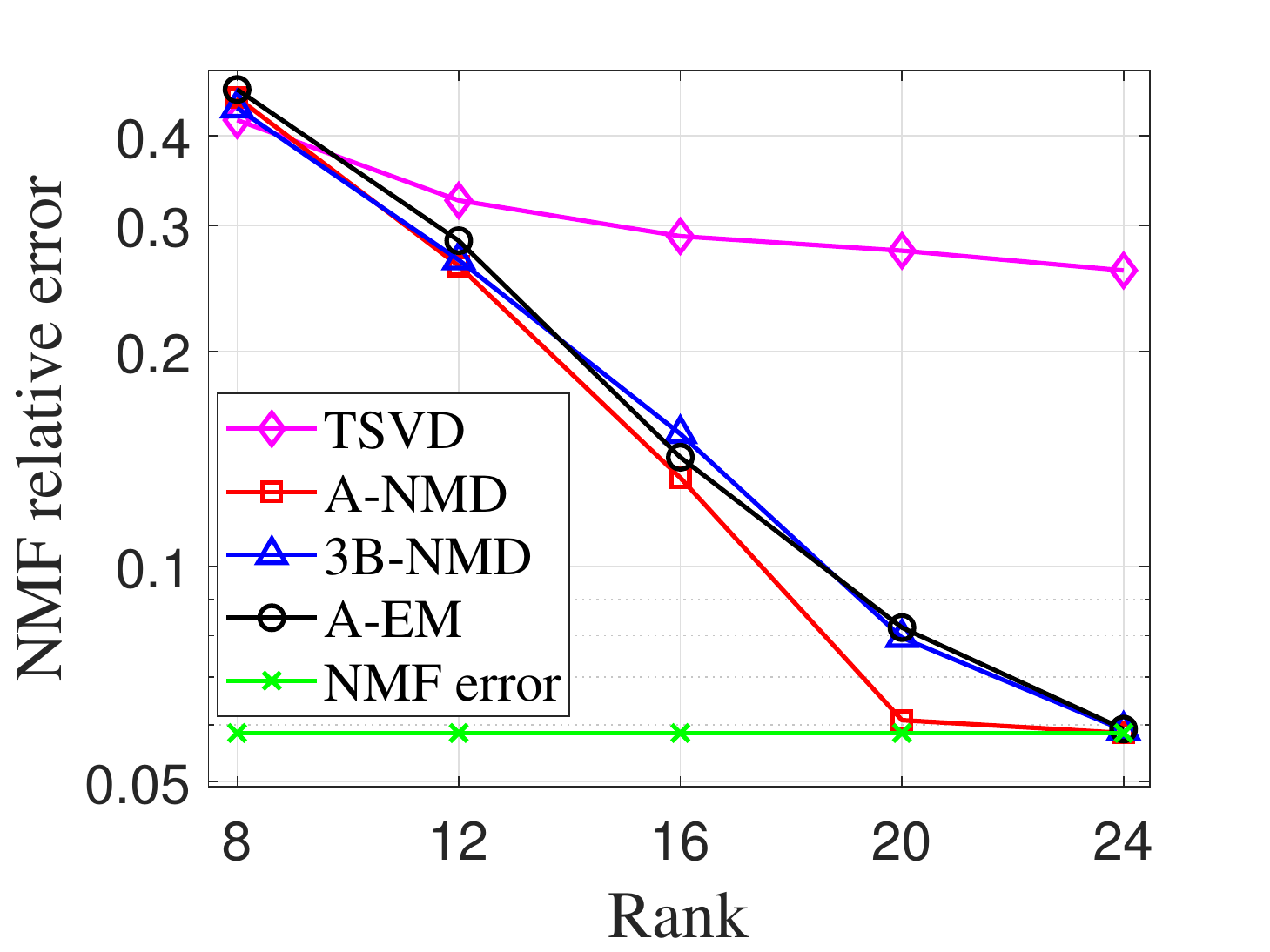}}
  \centerline{(b) Error on $X$; see~\eqref{eq:err_NMF}.}\medskip
\end{minipage}
\caption{Compression of a 361-by-100 NMF basis, $U$, of the CBCL data set. 
Image~(a) shows the error 
on the NMF basis $U \geq 0$. 
Image~(b) shows the NMF error after $U$ is replaced by its approximation; see~\eqref{eq:err_NMF}.}
\label{fig:rank_CBCL}
\end{figure}

\begin{figure}[htb]

\begin{minipage}[b]{.48\linewidth}
  \centering
  \centerline{\includegraphics[width=3.5cm]{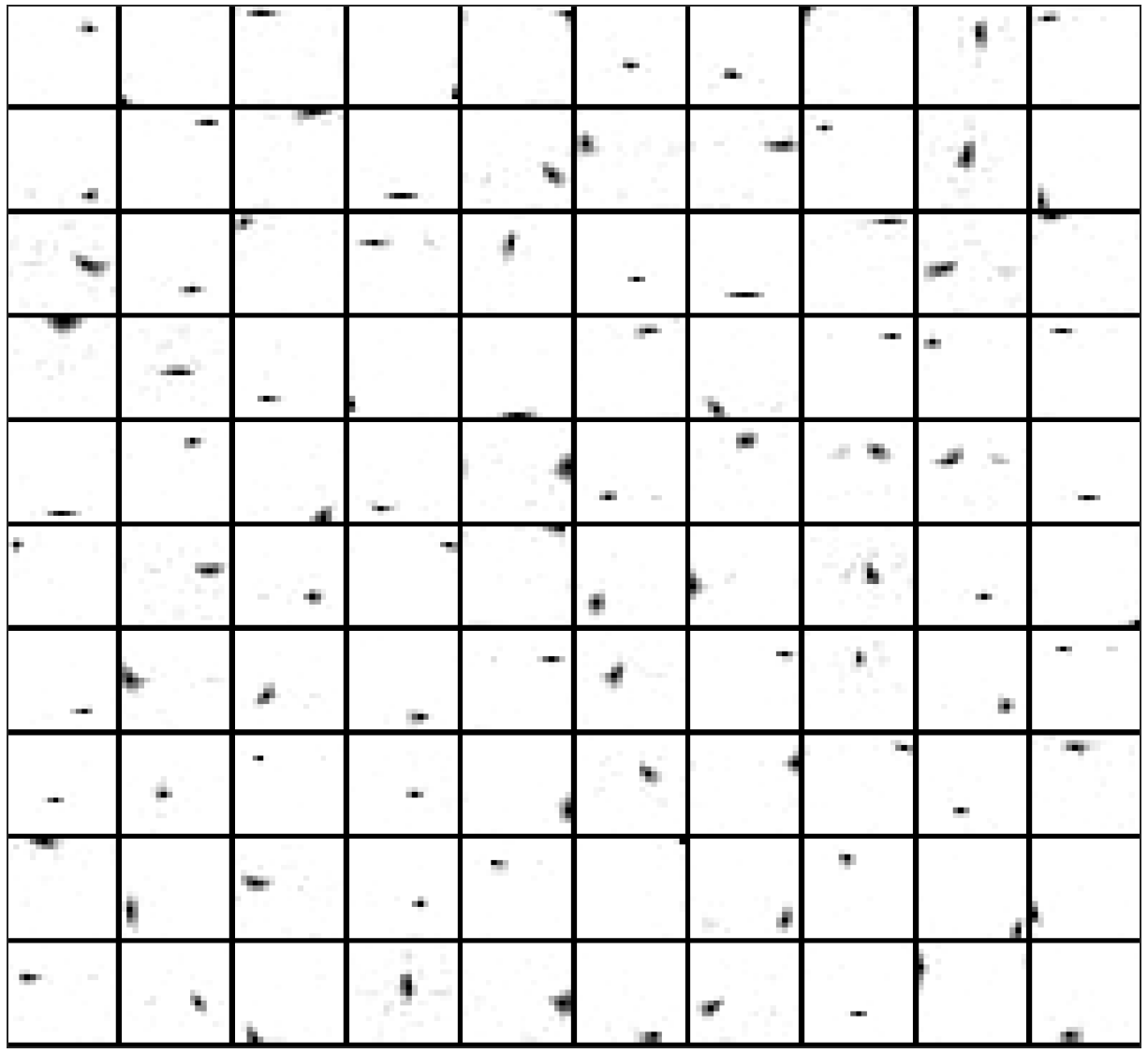}}
  \centerline{(a) Original $r=100$}\medskip
\end{minipage}
\hfill
\begin{minipage}[b]{0.48\linewidth}
  \centering
  \centerline{\includegraphics[width=3.5cm]{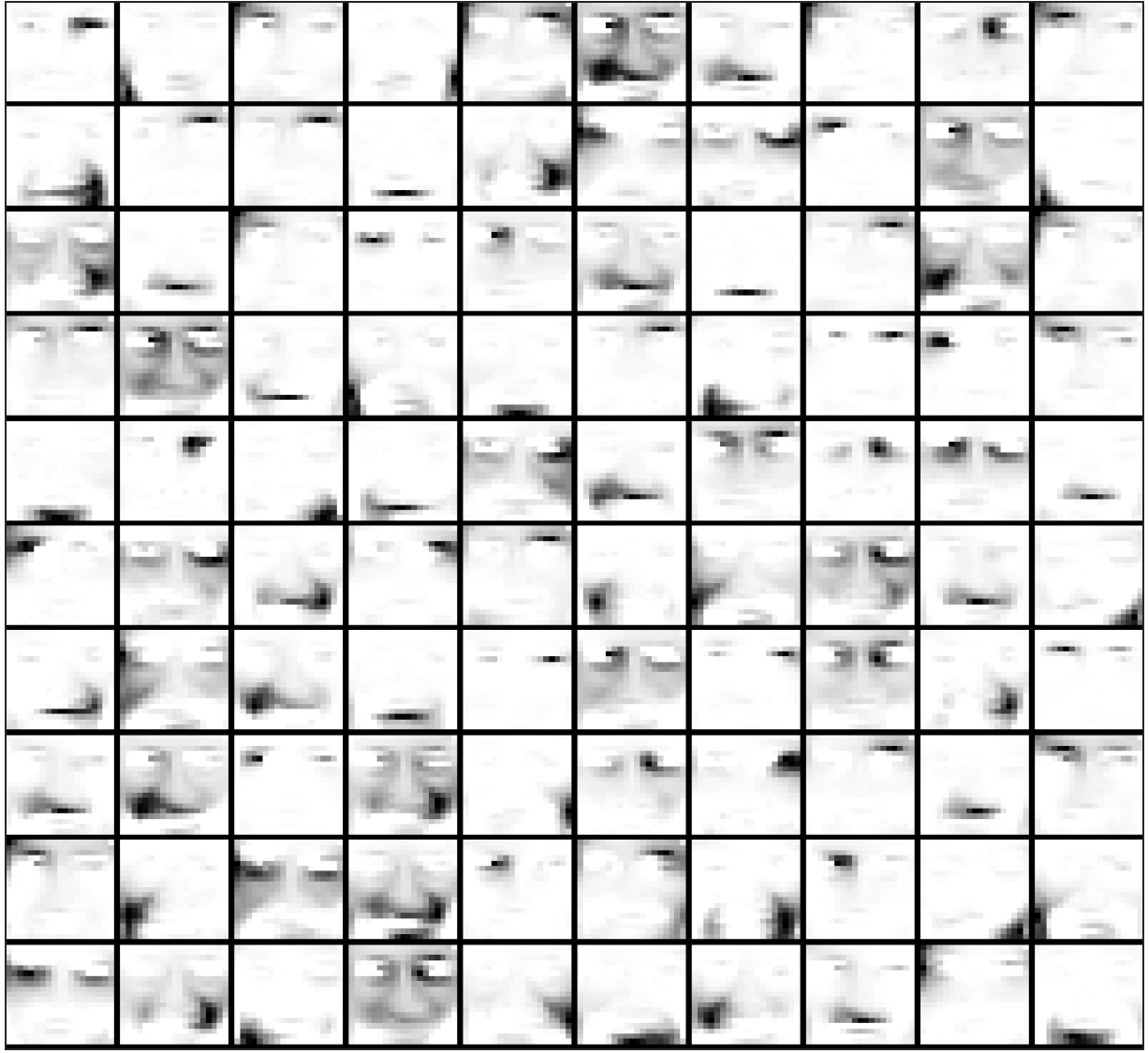}}
  \centerline{(b) TSVD $r=20$}\medskip
\end{minipage}
\begin{minipage}[b]{.48\linewidth}
  \centering
  \centerline{\includegraphics[width=3.5cm]{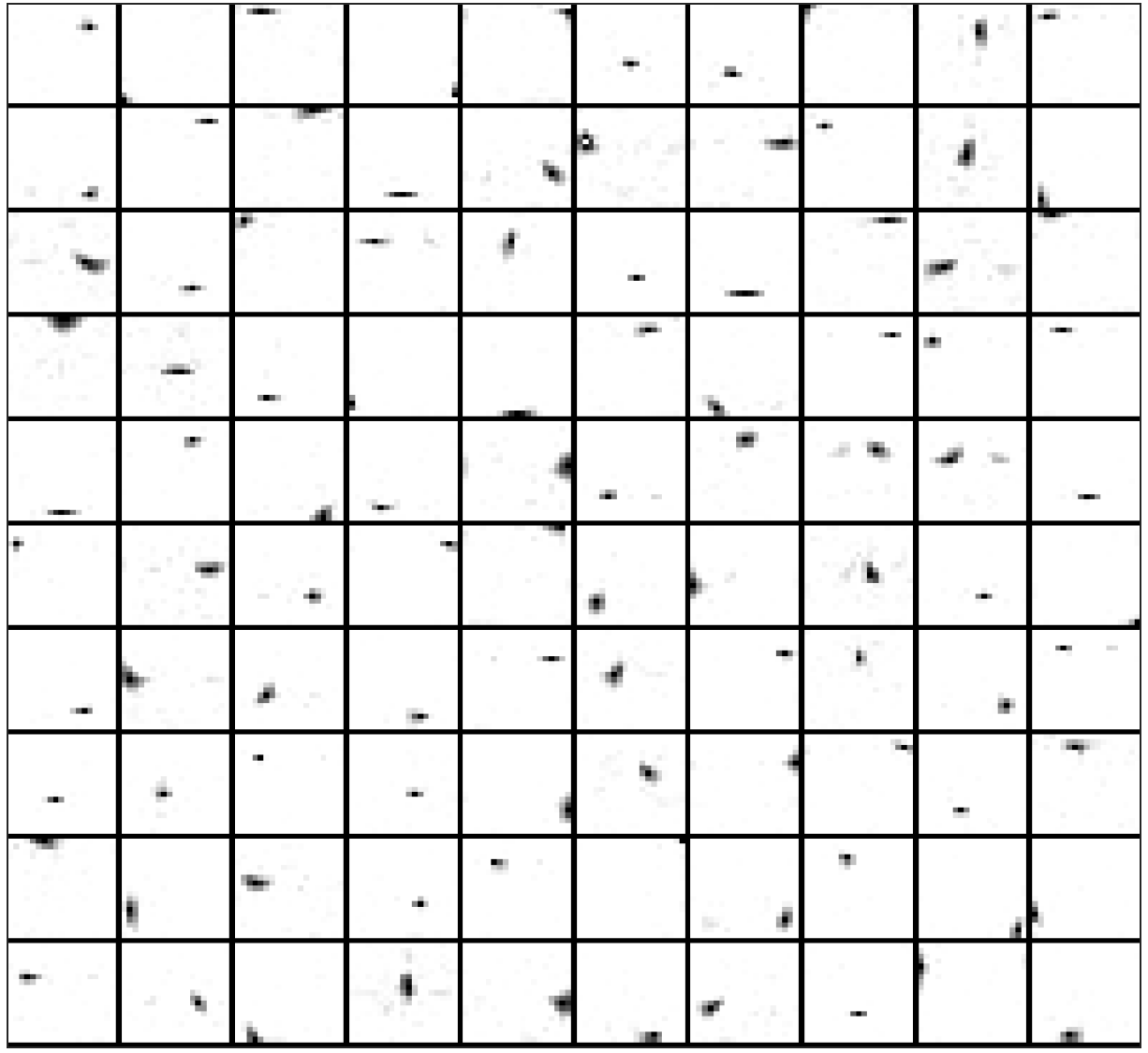}}
  \centerline{(c) A-NMD $r=20$}\medskip
\end{minipage}
\hfill
\begin{minipage}[b]{0.48\linewidth}
  \centering
  \centerline{\includegraphics[width=3.5cm]{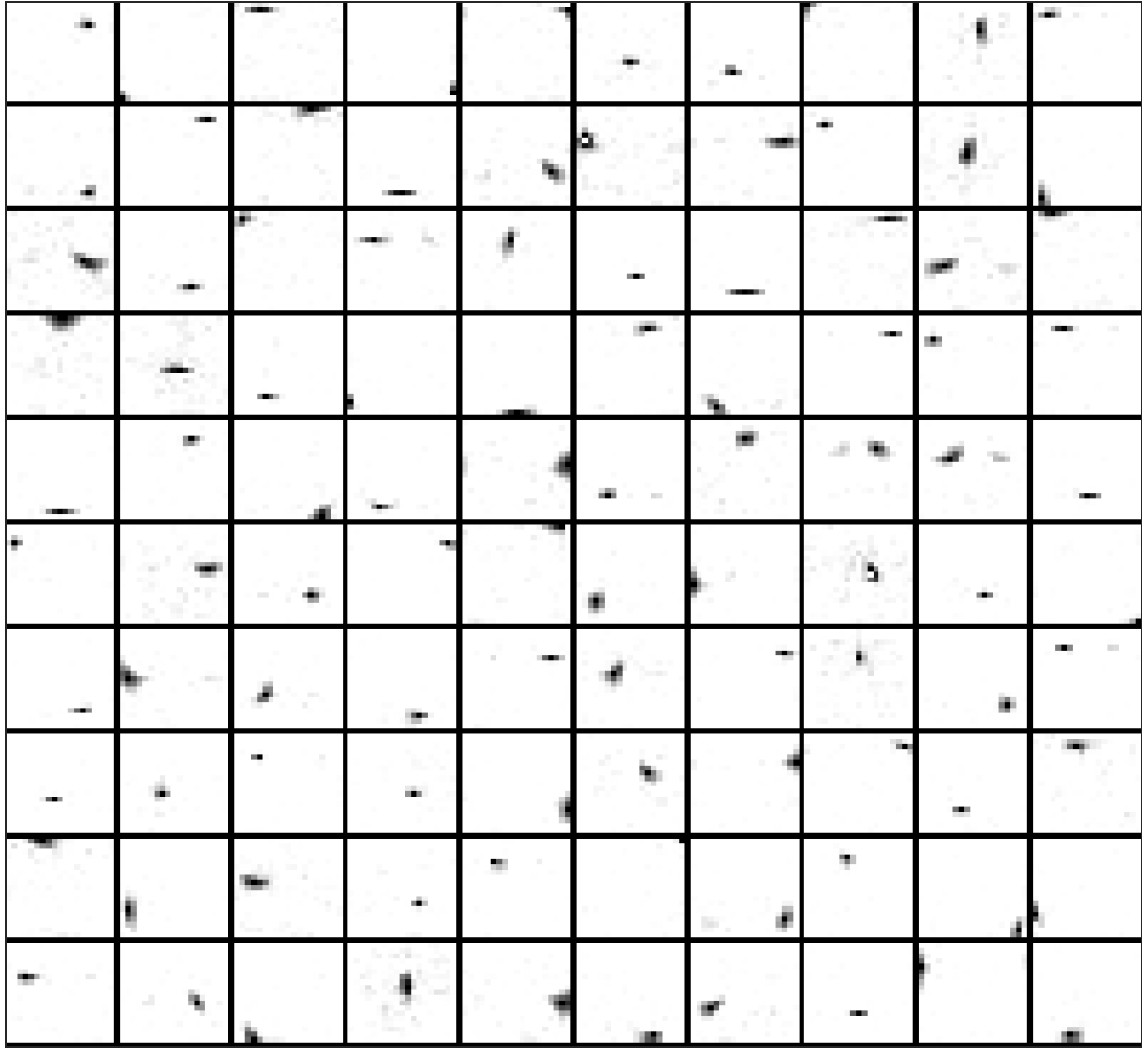}}
  \centerline{(d) 3B-NMD $r=20$}\medskip
\end{minipage}
\caption{Original factor $U$ of NMF, with rank-$r$=100 and low rank reconstruction by TSVD, A-NMD, and 3B-NMD with fixed rank-$r$=24. }
\label{fig:cbcl_rec}
\end{figure}

Another application of ReLU-NMD, which is new to the best of our knowledge, is the compression of sparse nonnegative dictionaries, e.g., the factors generated by NMF. 
Let us illustrate this on the CBCL data set, used in the seminal paper of Lee and Seung \cite{lee1999learning}. 
Each column of the data matrix $X \in \mathbb{R}^{361 \times 2429}$ contains a vectorized facial image of size $19 \times 19$. The NMF decomposition, $X \approx UV$ where $U \geq 0$ and $V \geq 0$, allows one to extract sparse facial features as the columns of $U$. To do so, we have used the regularized minimum-volume NMF code from \url{gitlab.com/ngillis/nmfbook/}, and used a rank-100 NMF to obtain 
$U \in \mathbb{R}^{361 \times 100}$, a nonnegative sparse matrix (in fact, 85\% of the entries are equal to zero); see Figure~\ref{fig:cbcl_rec}~(a) for an illustration. Further compressing the NMF factor $U$ using the TSVD is not  effective as Figure~\ref{fig:rank_CBCL}~(a) shows, with a relative error larger than 70\%  see also Figure~\ref{fig:cbcl_rec}~(b).  
This is because NMF tends to generate factors $U$ whose singular values are all large. However, ReLU-NMD does not have this limitation: it can approximate well such sparse full-rank matrices (e.g., the identity matrix~\cite{saul2022nonlinear}). 
Denoting $\hat{U} = \max(0,\Theta)$ the approximation matrix obtained by a ReLU-NMD algorithm, we first evaluate the relative error as in (\ref{eq:rel_err_std}) between $U$ and $\hat{U}$. Results are displayed for increasing values of the 
rank in Figure~\ref{fig:rank_CBCL}~(a). 
We also evaluate the error of the compressed NMF
\begin{equation}
    e_{NMF} = \min_{\hat{V} \geq 0} \frac{\rVert X - \max(0,\hat{U}) \hat{V}  \lVert_F }{\rVert X \lVert_F}, 
    \label{eq:err_NMF}
\end{equation}
see Figure~\ref{fig:rank_CBCL}~(b). 
In these experiments, we fix a time limit of 20 seconds. 
For such data sets, it appears that A-NMD reaches the best solution. For example, with $r=20$, it is able to reach almost the same accuracy on the NMF problem as the original factor of size $100$. 
Figure \ref{fig:cbcl_rec} shows an example of a rank $r=20$ reconstruction of the original rank $r=100$ NMF factor. 

\section{Conclusion}  
In this paper, we have proposed new algorithms for ReLU-NMD, one accelerating the naive algorithm by Saul (A-NMD), the other reparametrizing the low-rank variable to reduce the computational cost (3B-NMD). We showed their efficiency compared to the state of the art on synthetic and real-world data sets. 

\bibliographystyle{ieeetr}
\bibliography{References.bib}

\end{document}